\documentclass[a4paper]{svproc}
\usepackage{url}
\usepackage[numbers]{natbib} 
\usepackage{subcaption}
\usepackage{graphicx,wrapfig}
\usepackage[dvipsnames]{xcolor}

\begin{document}
\mainmatter              
\title{Constrained Robotic Navigation on Preferred Terrains Using LLMs and Speech Instruction: Exploiting the Power of Adverbs}
\titlerunning{Offroad Navigation based on Language models}  
%
\author{Faraz Lotfi \and Farnoosh Faraji \and Nikhil Kakodkar \and  Travis Manderson \and 
David Meger \and Gregory Dudek }
\authorrunning{Faraz Lotfi et al.} 
%
%
\institute{MRL Lab, McGill University, Montreal, Canada,\\
\email{{F.Lotfi, Farnoosh, nikhil, travism, DMeger, Dudek}@cim.mcgill.ca}}

\maketitle              

%
\section{Abstract}
This paper explores leveraging large language models for map-free off-road navigation using generative AI, reducing the need for traditional data collection and annotation. We propose a method where a robot receives verbal instructions, converted to text through Whisper, and a large language model (LLM) model extracts landmarks, preferred terrains, and crucial adverbs translated into speed settings for constrained navigation. A
language-driven semantic segmentation model generates text-based masks for identifying  landmarks and terrain types in images. By translating 2D image points to the vehicle's motion plane using camera parameters, an MPC controller can guides the vehicle towards the desired terrain. This approach enhances adaptation to diverse environments and facilitates the use of high-level instructions for navigating complex and challenging terrains.
\keywords{Constrained map-free navigation, large language models, language-driven semantic segmentation, preferred terrains, speech instruction, adverbs.}

\section{Introduction}
    In this paper, we examine the fundamental problem of incorporating broad prior knowledge into robotics planning problems.  We do this by exploiting large language models which are able to encode diverse priors.  This builds on our previous work where we exploited the generality of such models to encode real world contextual constraints for other domains of operation~\cite{greg}.
    
\begin{wrapfigure}{r}{0.27\textwidth}
    \includegraphics[width=0.25\textwidth]{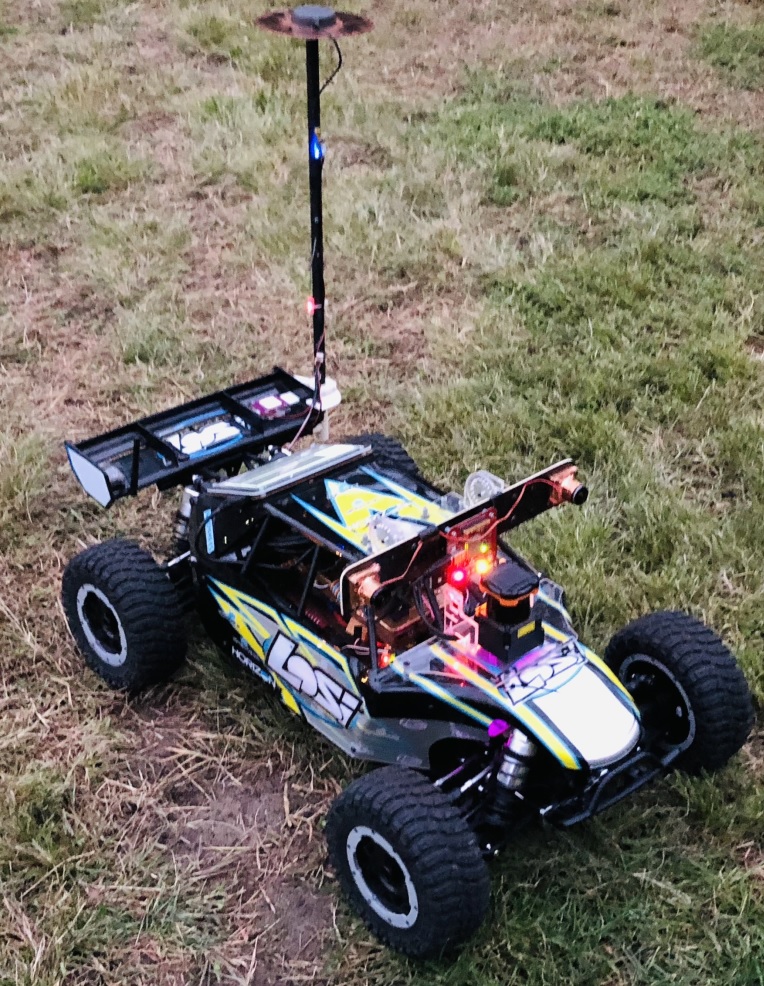}\\
    \caption{The intended offroad vehicle}
    \label{fig:rc_car}
\end{wrapfigure}

In the context of outdoor navigation tasks involving off-road vehicles, such as the one depicted in Fig. \ref{fig:rc_car}, individuals often guide a robot towards terrains with other observable properties (e.g. smoothness, or dry, or packed earth). This can be accomplished by employing predictive models that use current observations to suggest a series of actions, guiding the robot towards the desired ground conditions \cite{badgr}. To achieve this, researchers typically collect data from real-world environments and subsequently annotate the data either automatically or manually, treating the planning problem as a classification task at each time step. The goal is to generate a set of actions based on the current observation and the corresponding predicted events. 

Several models have been previously explored such as visually learning to drive on smooth terrain by combining IMU data with onboard and aerial images ~\cite{manderson2020-icra} or using trajectory constrained attention models to improve learning performance~\cite{wapnick2021}. These approaches, however, suffer from various limitations.  Generalization is a common challenge, so that when attempting to navigate the robot in a relatively new environment, new data must be gathered. Additionally, the annotation process is burdensome. Automatically annotating  data through the use of IMU or Lidar measurements 
can sometims be effective, but can introduce noise into the dataset, as threshold values must be established without a standardized procedure. Manual annotation is time-consuming, posing challenges in terms of timeliness and distribution shifts when gathering new data.

On the other hand, significant progress has been made in the development of transformers, vision-language models (VLMs) and foundation models to leverage prior knowledge and
reduce the need for extensive annotation. One such model is CLIP~\cite{clip}, which incorporates two input streams for image and text embedding, and can be used to
derive a similarity metric. CLIP has found applications in robotics, even in our own work, such as assisting in automated photography in locating desired photos from recorded videos of special events~\cite{greg}, or enabling object grasping without the need for explicit target detection by an object detector~\cite{icra2023}.

Another model in this field is LSeg~\cite{lseg}, which shares similarities with CLIP as it also uses two input streams for text and image embedding, but outputs text grounded at the pixel level. Unlike CLIP, which approaches the problem from a classification perspective, LSeg matches the queried text with each pixel embedding, thereby formulating the problem as a segmentation task.

To gain further leverage from these models, researchers have combined them with large language models~\cite{greg, lmnav, vlmaps}. These combined models can be employed in impressive ways for planning purposes, analyzing human-provided instructions to extract landmarks or spatial goals. For example, LM-nav~\cite{lmnav} utilizes large language models for outdoor navigation by extracting landmarks from text instructions, while Vlmaps~\cite{vlmaps} adds the capability of spatial goal inquiries to the instructions, enabling navigation in indoor environments.

By combining the strengths of several types of language-based model, we propose an approach that goes beyond sparse information such as landmarks or spatial goals, enabling the inclusion of rich descriptors in instructions. In this paper, we examine the use of high-level verbal cues to robots, which appear to allow for both more intuitive and flexible specifications than traditional methods. Our approach incorporates constraints in the form of adverbs within the instructions and also includes orders for selecting different terrains over time.

As our second contribution, we use LSeg to achieve language-driven semantic segmentation when presented with various terrains eliminating the need for annotations or gathering new data. The semantic segmentation map is then used by a local planner to guide the robot toward the desired terrains while accomplishing a set of goals. This not only leads to a more generalized planner but also allows the robot to handle unforeseen challenges. For example, the user may be aware that a smooth road in the intended environment contains undetectable holes, which can lead to failures. In such cases, the desired terrain specified in the instruction could be an alternative, such as an existing traversable grass or moving slowly on the smooth road.

It is worth noting that most existing work on planning assume prior exploration of the environment, resulting in a representation of the environment in the form of a grid map or a graph. This representation facilitates navigation and goal achievement. In this article, we explore planning solely based on the high-level instructions, without relying on any prior map.
\section{Methodology}
\begin{figure}[h]
	\begin{center}
	   \includegraphics[width=1\linewidth]{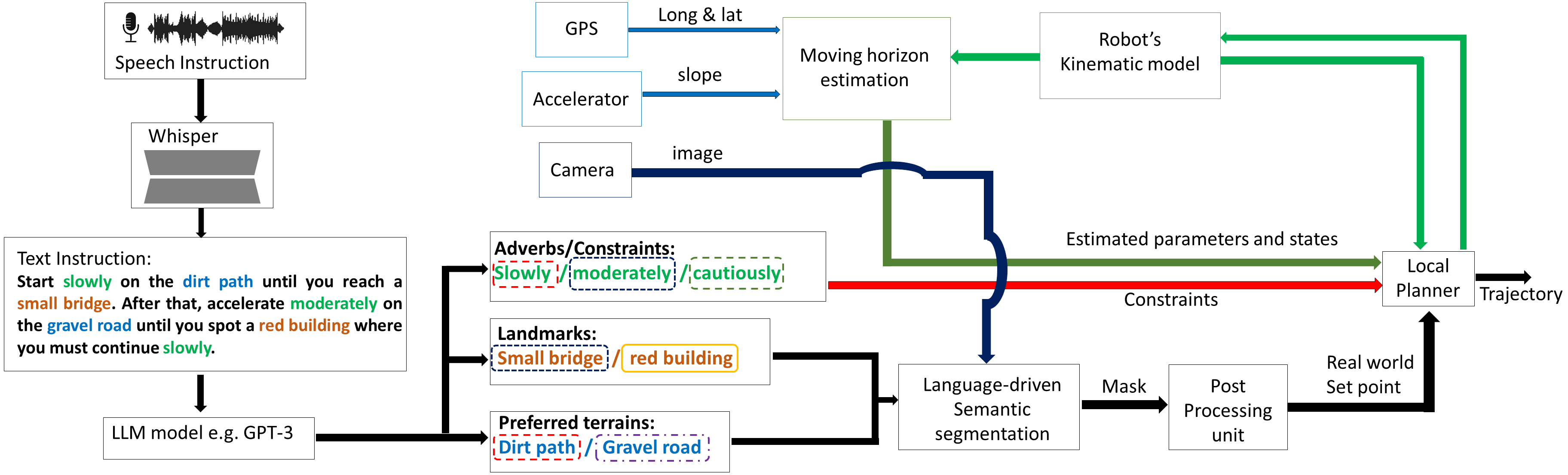}
	\end{center}
	\caption{The overall block diagram of the proposed navigation system}
	\label{fig:paradigm}
\end{figure}

The block diagram illustrating the entire planner paradigm can be seen in Fig. \ref{fig:paradigm}. In what follows, we will delve into the specifics of each module.\\
    \textbf{- Speech-to-text model:}\\
    This module is responsible for converting spoken instructions into text format, which can then be utilized by the LLM module. To accomplish this, we employ whisper \cite{whisper}, utilizing a zero-shot approach that achieves high accuracy in converting spoken instructions to text.\\    
    \textbf{- LLM module:}\\
    There has been a recent surge in intest in Large language models in robotics applications. Numerous LLMs have been proposed in the literature, including~\cite{gpt-j,gpt-neox20b,gpt3}, each developed with varying numbers of parameters. Among these models, we select GPT-3~\cite{gpt3}, which has demonstrated successful usage in~\cite{greg,icra2023,lmnav,vlmaps} through the zero-shot approach. We have observed that GPT-3.5 effectively extracts the pertinent contextual information from the given instruction, such as landmarks, preferred terrains, and adverbs as constraints. Compared to existing literature, we augment the given instruction with adverbs and possible terrain types.\\
    \begin{figure}[]
	\begin{center}
	   \includegraphics[width=0.9\linewidth]{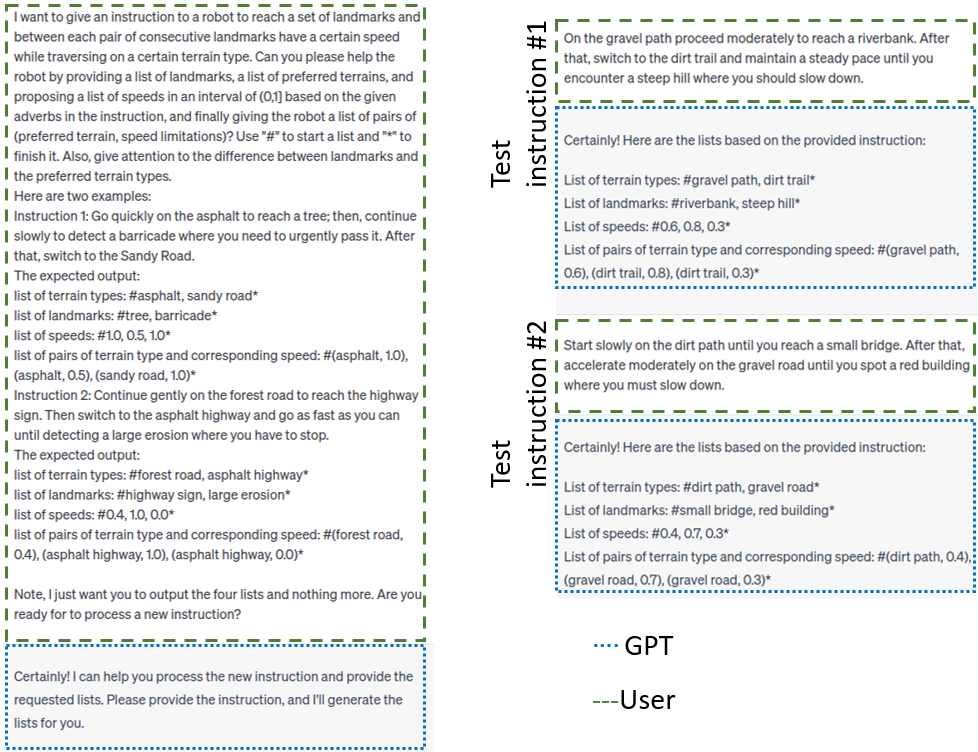}
	\end{center}
	\caption{Engineering an effective prompt to elicit the desired output from a Large Language Model (LLM).
    }
	\label{fig:prompt_eng}
    \end{figure}
    To effectively harness the power of an LLM such as GPT-3, prompt engineering is essential. This module's output is utilized downstream by other modules, and prompt engineering ensures the generation of precise output tailored for use by these downstream components. Fig. \ref{fig:prompt_eng} illustrates the instructions provided to the LLM. It's important to emphasize that offering a specific example of the desired outcome greatly improves the quality of the results. Furthermore, we firmly believe that increasing the number of examples would contribute to achieving superior outcomes and better adapting to potential instructions. Moreover, commencing lists with distinctive markers such as ``$\star$'' and/or ``$\#$'' enables us to efficiently extract essential information while processing such outputs.\\
     By employing a list that associates terrain types with their corresponding speeds and a separate list featuring landmarks, the robot attains autonomous navigation, smoothly transitioning between different terrains or speeds with the aid of landmarks. It's crucial to emphasize that this approach relies on the assumption that landmarks are readily available to facilitate switches between terrains, speeds, or a combination of both. Lastly, these models operate without any prior training. \\
    \textbf{- Language-driven semantic segmentation}\\
   We equip language-driven models with a list of queries that encompass preferred terrains and landmarks. This empowers the model to perform both terrain type and landmark detection. The outcome of this process is a grounded text segmentation map, which, in turn, serves as a basis for situation inference and the generation of a desired goal location for the local planner.\\
   In order to extract the essential information from the provided segmentation map, a post-processing step is essential to ensure robust map inference. To achieve this, we employ average pooling with varying kernel sizes, starting with larger ones, to identify the predominant pixel class within different sections of the image. By analyzing the class scores assigned to each pixel, we can effectively eliminate noise and areas of uncertainty within the image, resulting in a reliable outcome. \\
   \textbf{- Local planner}\\
    For the local planner, we adopt a nonlinear Model Predictive Control (MPC) approach in conjunction with a kinematic model of the robot~\cite{rcmodel}. An advantageous aspect of MPC is the ability to incorporate adverbs into the planning process as constraints that must be satisfied. \\ 
    \textbf{- Moving horizon estimator (MHE)}\\
    MHE~\cite{MHE} estimates the model's states and parameters simultaneously, enabling the system to handle uncertainties arising from both the model and the environment.
 
\section{Results}
In this section, we present an evaluation of language-driven segmentation models, specifically LSeg and Conceptfusion~\cite{conceptfusion} using two distinct datasets. 

We began our evaluation by testing the performance of these models using the dataset introduced in~\cite{dataset1}. Specifically, we provided the models with a set of queries, "other" and "road." The dataset in question contains a total of 5,379 images, each accompanied by its respective attention regions. To assess the results, we utilized the dice metric~\cite{dice}, a well-established measure for evaluating segmentation maps. Furthermore, in order to gain deeper insights into the models' performances, we partitioned the dataset into 10 fragments based on the ratio of true segment coverage relative to the entire image. This fragmentation allowed us to conduct a better analysis of how effectively each model handles various true segment sizes. Subsequently, we calculated both the mean and variance of the dice values, as these two metrics are crucial for a comprehensive assessment.

\begin{figure}[h]
    \centering
    \begin{subfigure}[b]{\textwidth}
        \centering
        \includegraphics[width=1.0\textwidth]{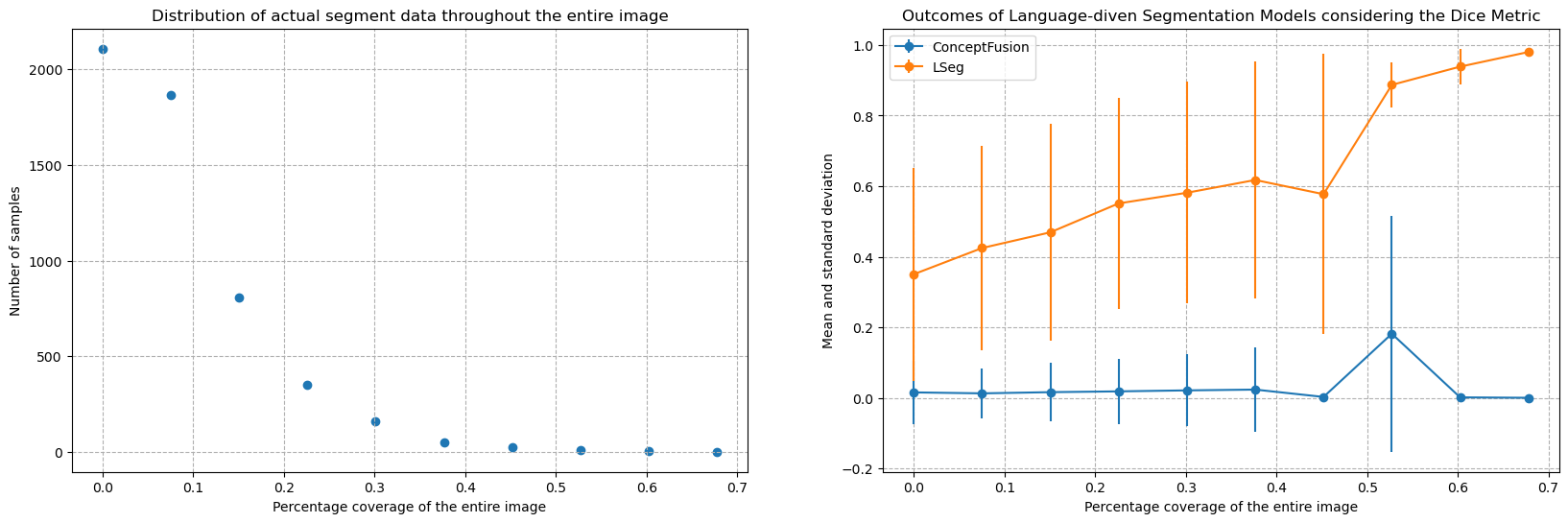}%
    \end{subfigure}
    \caption{This figure depicts the performance of two different models on an unseen dataset of offroad scenes. The left plot illustrates the distribution of data based on the percentage of coverage of the truth segment throughout the entire image, while the right plot displays the results for the dice metric. As evident from the results, LSeg demonstrates improved performance with larger regions of interest, while Conceptfusion appears to struggle in this particular application. }
    \label{fig:results_dataset1}
\end{figure}

\begin{figure}[h]
    \centering
    \begin{subfigure}[b]{\textwidth}
        \centering
        \includegraphics[width=1.0\textwidth]{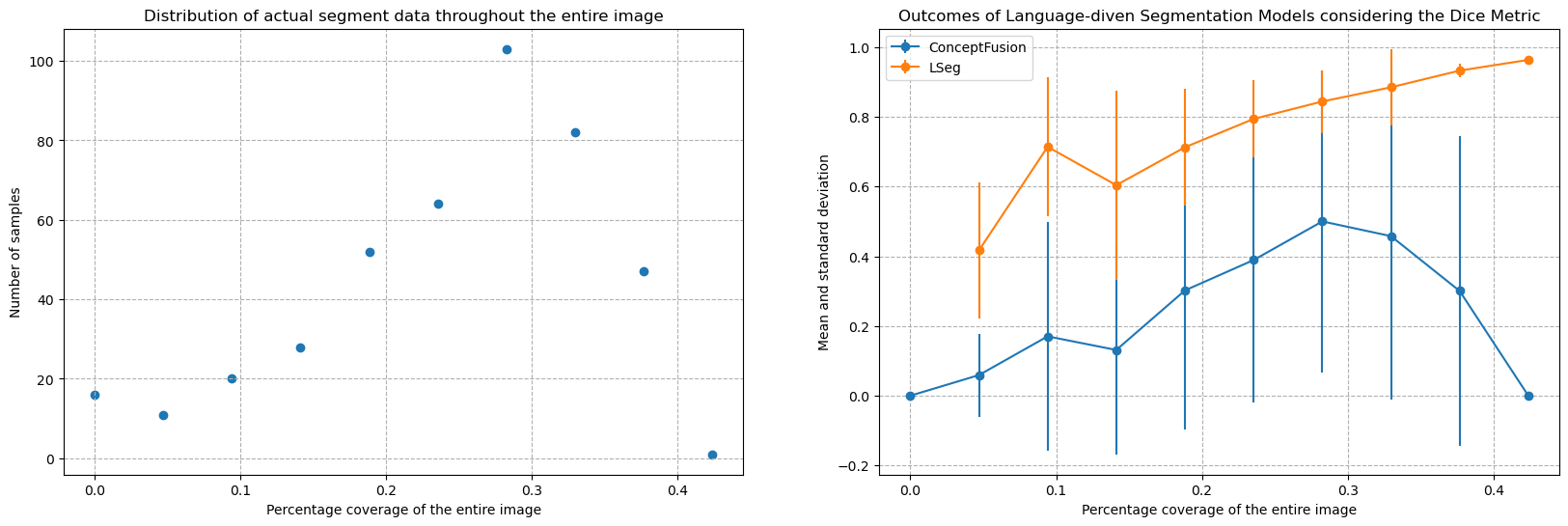}%
    \end{subfigure}
    \caption{These plots depict the results obtained from evaluating the models on a dataset that features more samples with larger true segments. Crucially, this dataset was acquired using a front-facing camera mounted on our RC car, navigating through challenging off-road terrain.}
    \label{fig:results_dataset2}
\end{figure}

Fig. \ref{fig:results_dataset1} presents the results obtained in our study. The LSeg model displayed promising performance when dealing with segments covering more than $20\%$ of the image; however, its performance declined for smaller segments. Despite this limitation, its impressive performance on larger segments suggests its potential for extracting features like terrain types. Significantly, among the models, LSeg stands out as the superior choice for applications relative to offroad navigation, while Conceptfusion demonstrates poor performance in this case.

\begin{figure}[h]
	\begin{center}
	   \includegraphics[width=1\linewidth]{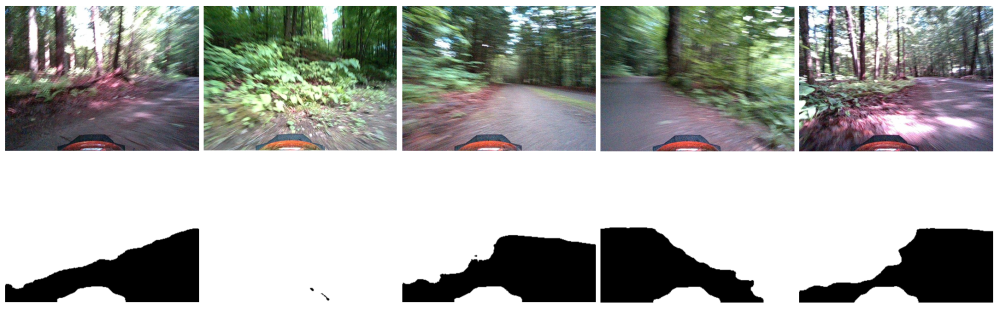}
	\end{center}
	\caption{Samples from the second dataset, paired with masks generated by LSeg, provide insight into the evident pathways for navigation. The interesting observation is the model's impressive ability to effectively filter out the vehicle's front portion and produce minimal output in scenarios where no clear path exists.}
	\label{fig:exp0}
    \end{figure}

As shown in Fig. \ref{fig:results_dataset1}, the first dataset has a limited number of samples with significant true segment coverage, making it unsuitable for comprehensive model evaluation. To address this, we introduced a second dataset, derived from a separate project, featuring robot-mounted camera images with prominent road features and large true segments. Fig. \ref{fig:exp0} provides visual representations of segmentation maps generated by the LSeg model for samples from this dataset, demonstrating promising results that support the effective use of these models in scenarios with substantial true segments.

Fig. \ref{fig:results_dataset2} presents the results obtained when employing the two models on the second dataset. These results reaffirm our initial expectation, as seen in Fig. \ref{fig:results_dataset1}, that these models exhibit superior performance for larger segments. Furthermore, this finding underscores the importance of selecting the right model for our off-road navigation task, with the suitable choice being LSeg.

\begin{figure}[h]
	\begin{center}
	   \includegraphics[width=1\linewidth]{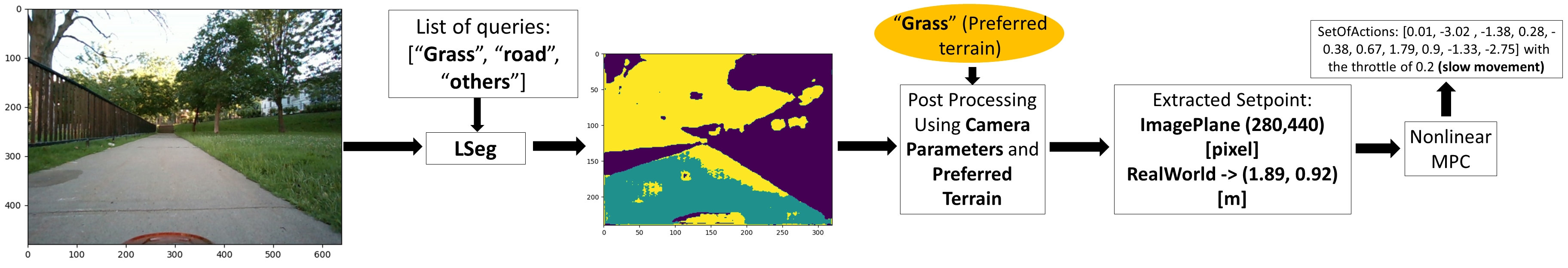}
	\end{center}
	\caption{An overview of the planner functioning}
	\label{fig:exp1}
    \end{figure}
\section{Experimental results}

We implemented the proposed scheme on a real-world mid-sized RC car, as illustrated in Fig. \ref{fig:rc_car}. This RC car is widely utilized as a ground robot for off-road navigation on smooth roads~\cite{manderson2020-icra,wapnick2021}. 
To leverage existing services for utilizing Large Language Models (LLMs), we employ the provided APIs. In the beginning of an experiment, we assume internet access is available to facilitate sending requests for both text generation and LLM model usage. It's important to note that this internet access is a prerequisite during the initial phase when the robot receives instructions. Subsequently, during the planning and execution of tasks, there is no requirement to re-utilize these two models, and therefore, no internet connection is needed for their continued use.
In contrast, we run the language-driven semantic segmentation model onboard continuously. This is essential for processing incoming images and providing the controller with the necessary inputs.

In Fig. \ref{fig:exp1}, we present an overview of the navigation system's operation. This process involves starting with the current image and extracting information using the LLM, which then feeds into the LSeg module to generate semantic segmentation outputs. Following this, depth estimation is employed to determine the necessary real-world setpoints, which are subsequently utilized by the MPC controller to generate the robot's actions.
\begin{figure}[h]
	\begin{center}
	   \includegraphics[width=1\linewidth]{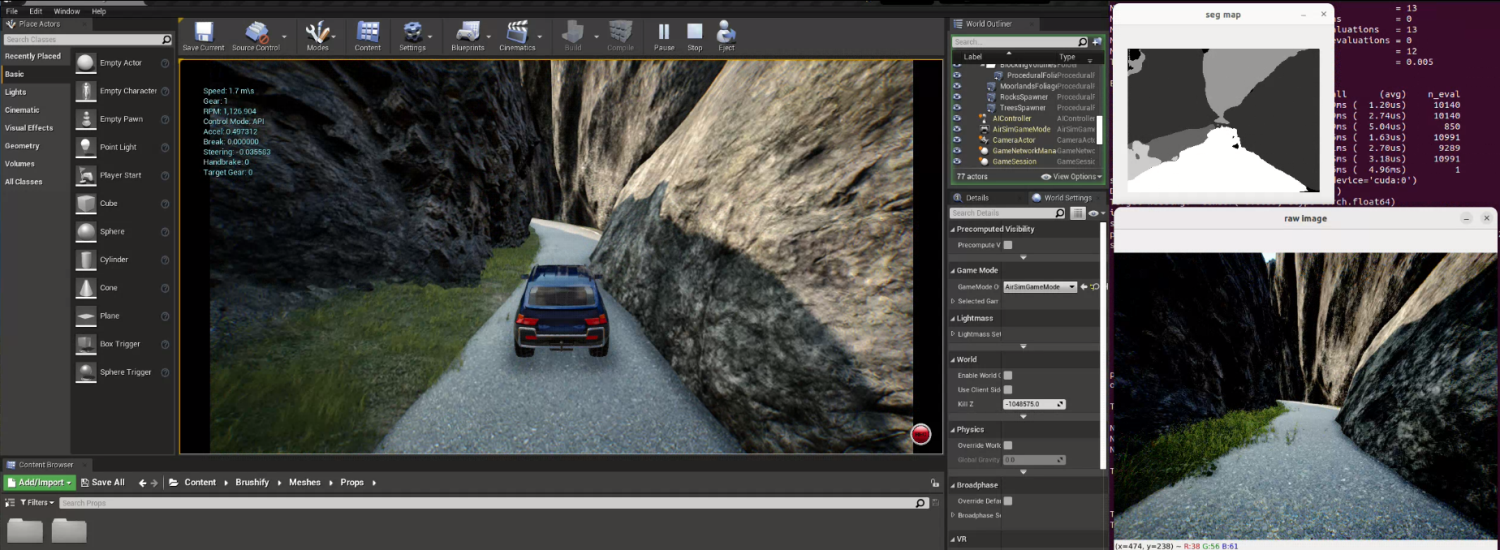}
	\end{center}
	\caption{This figure displays our experimental environment. On the left is a third car view, and on the right, the front-view camera output with LSeg model's semantic segmentation.}
	\label{fig:unreal_env}
    \end{figure}
\begin{figure}[h]
	\begin{center}
	   \includegraphics[width=1\linewidth]{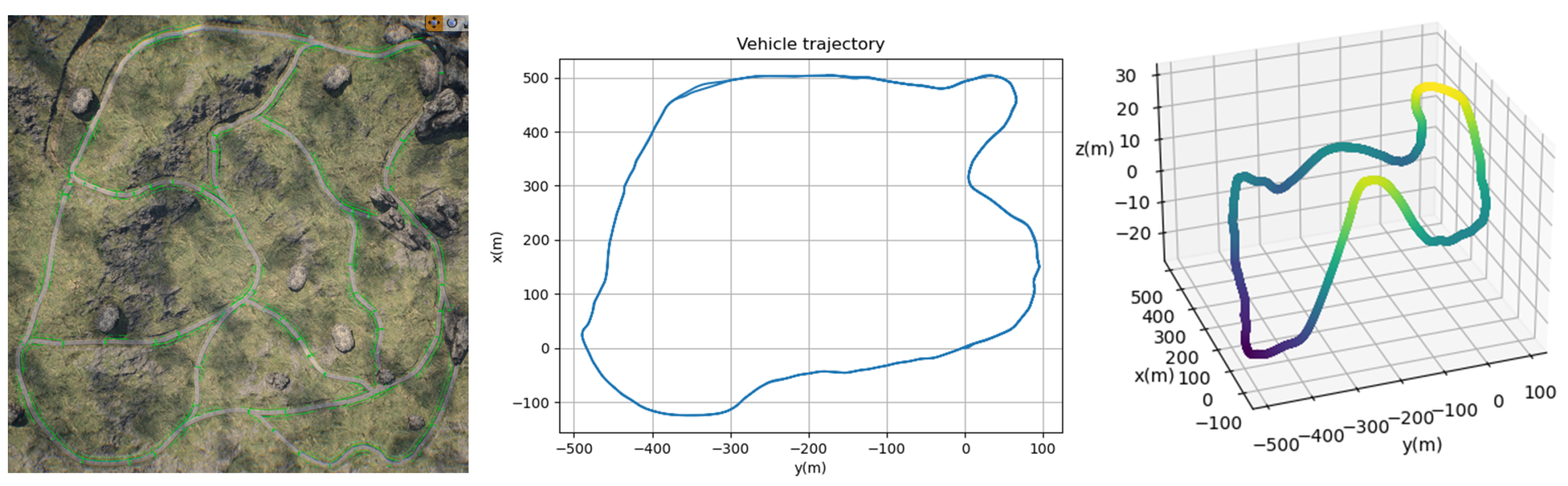}
	\end{center}
	\caption{In our first study, the trajectory encompassed extended periods of successful navigation in a vast and challenging terrain. The figure includes a top-view of the environment, along with 2D and 3D traversed trajectories, moving from left to right.  }
	\label{fig:unreal_results}
    \end{figure}
\begin{figure}[h]
	\begin{center}
	   \includegraphics[width=1\linewidth]{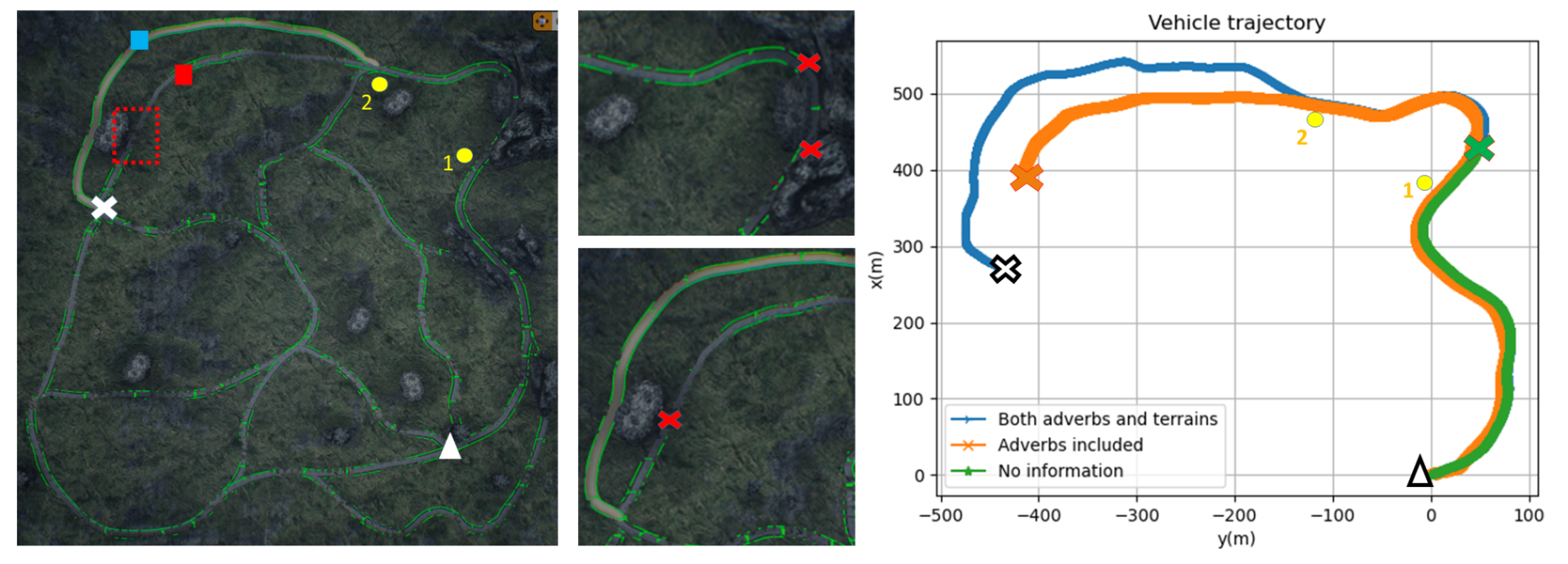}
	\end{center}
	\caption{In the second study, we consider two key landmarks denoted by yellow circles, where speed, terrain preference, or both are altered. On the left, the environmental map is displayed, featuring a white triangle and cross representing the start and end points, respectively. A solid red box signifies a blocked path, while a blue box represents a clear path with a different terrain type. Additionally, a red-dotted bounding box highlights a few obstacles obstructing the route. In the middle section, the upper part reveals two high-risk locations where improper speed can lead to failures, while the lower part illustrates another potential failure scenario due to obstacles. The final plot demonstrates the performance of a planner with different information, underscoring the importance of providing comprehensive instructions in reaching a goal successfully despite the challenges.     }
	\label{fig:unreal_study2}
    \end{figure}    
For the sake of reproducibility, we conducted our extensive experiments within the Unreal game engine environment, where we initially evaluated the long-term navigation performance of the proposed paradigm. Fig. \ref{fig:unreal_env} offers a glimpse of our setup, showcasing the test environment. By running our pipeline in this environment for extended durations, we ensure the pipeline's stability.

Fig. \ref{fig:unreal_results} illustrates the results in the form of a 3D trajectory, showcasing the traversed path. Notably, the environment includes slopes, curved roads, obstacles, and challenging scenarios, as exemplified in Fig. \ref{fig:unreal_env}. The planner effectively maintains the vehicle on the road, demonstrating consistent performance without divergence or fluctuations during multiple runs along this extensive route. It's worth noting that the vehicle successfully navigated the same path multiple times, accumulating hours of navigation without any failures.

In our latest experiment, we conducted a trial in which we provided specific instructions to the robot, including landmarks and preferred terrains, to assess its ability to successfully navigate while avoiding blocked paths and maintaining a safe speed. The instructions allowed us to control the robot's speed effectively, especially in challenging situations where excessive speed could result in failures.

This study involved two key landmarks: a parked car and an animal sculpture. We also considered two distinct terrains: asphalt and sandy roads. The scenario began with a brisk start at 3 m/s. Upon reaching the parked car, the robot's speed was automatically reduced to 1.5 m/s due to the presence of a mountainous region with curves, slopes, and a narrow path. Subsequently, after detecting the second landmark, the preferred terrain transitioned to a sandy road, as the asphalt road was blocked further ahead. 

Fig. \ref{fig:unreal_study2} provides a comprehensive summary of the results, showcasing a top-view map image. This visualization highlights various key elements, including blocked paths, landmarks, terrain types, and the initial and final points. To enhance the robustness of our findings, we performed multiple iterations of the simulation.

Furthermore, to gain deeper insights into the significance of incorporating adverbs and preferred terrain types in instructions, we conducted two additional studies. One study omitted adverbs, while the other omitted references to preferred terrains. The results, displayed within the same figure, clearly underscore the effectiveness of enriched instructions. It is evident that the absence of adverbs or preferred terrains resulted in navigation failures for the vehicle.

 \section{Conclusion}
In this study, we investigated the efficacy of providing speech instructions to a robot for off-road navigation, incorporating landmarks, preferred terrains, and adverbs to enhance guidance. Adverbs are pivotal in specifying the required speed on preferred terrains, while landmarks trigger terrain and speed adjustments. We employ Whisper to transcribe speech into text, followed by an LLM to extract the necessary information from the provided instruction. We then employ language-driven semantic segmentation maps and conduct a comparative analysis of two recent models using two datasets, demonstrating their strong generalization for off-road navigation. Practical deployment in a virtual environment confirms that enriching instructions with preferred terrains and adverbs significantly enhances performance, reducing failures.

%
%
%
\begingroup
\let\clearpage\relax
\bibliography{egbib.bib}

\begin{thebibliography}{18}
\providecommand{\natexlab}[1]{#1}
\providecommand{\url}[1]{{#1}}
\providecommand{\urlprefix}{URL }
\expandafter\ifx\csname urlstyle\endcsname\relax
  \providecommand{\doi}[1]{DOI~\discretionary{}{}{}#1}\else
  \providecommand{\doi}{DOI~\discretionary{}{}{}\begingroup
  \urlstyle{rm}\Url}\fi
\providecommand{\eprint}[2][]{\url{#2}}

\bibitem[{Rivkin et~al(2023)Rivkin, Dudek, Kakodkar, Meger, Limoyo, Jenkin,
  Liu, and Hogan}]{greg}
Rivkin D, Dudek G, Kakodkar N, Meger D, Limoyo O, Jenkin M, Liu X, Hogan F
  (2023) Ansel photobot: A robot event photographer with semantic intelligence.
  In: 2023 IEEE International Conference on Robotics and Automation (ICRA),
  IEEE, pp 8262--8268

\bibitem[{Kahn et~al(2021)Kahn, Abbeel, and Levine}]{badgr}
Kahn G, Abbeel P, Levine S (2021) Badgr: An autonomous self-supervised
  learning-based navigation system. IEEE Robotics and Automation Letters
  6(2):1312--1319

\bibitem[{Manderson et~al(2020)Manderson, Wapnick, Meger, and
  Dudek}]{manderson2020-icra}
Manderson T, Wapnick S, Meger D, Dudek G (2020) Learning to drive off road on
  smooth terrain in unstructured environments using an on-board camera and
  sparse aerial images. In: 2020 IEEE International Conference on Robotics and
  Automation (ICRA), IEEE, pp 1263--1269

\bibitem[{Wapnick et~al(2021)Wapnick, Manderson, Meger, and
  Dudek}]{wapnick2021}
Wapnick S, Manderson T, Meger D, Dudek G (2021) Trajectory-constrained deep
  latent visual attention for improved local planning in presence of
  heterogeneous terrain. In: 2021 IEEE/RSJ International Conference on
  Intelligent Robots and Systems (IROS), IEEE, pp 460--467

\bibitem[{Radford et~al(2021)Radford, Kim, Hallacy, Ramesh, Goh, Agarwal,
  Sastry, Askell, Mishkin, Clark et~al}]{clip}
Radford A, Kim JW, Hallacy C, Ramesh A, Goh G, Agarwal S, Sastry G, Askell A,
  Mishkin P, Clark J, et~al (2021) Learning transferable visual models from
  natural language supervision. In: International conference on machine
  learning, PMLR, pp 8748--8763

\bibitem[{Xu et~al(2023)Xu, Zhao, Zhou, Li, Pi, Zhu, Wang, and
  Xiong}]{icra2023}
Xu K, Zhao S, Zhou Z, Li Z, Pi H, Zhu Y, Wang Y, Xiong R (2023) A joint
  modeling of vision-language-action for target-oriented grasping in clutter.
  arXiv preprint arXiv:230212610

\bibitem[{Li et~al(2022)Li, Weinberger, Belongie, Koltun, and Ranftl}]{lseg}
Li B, Weinberger KQ, Belongie S, Koltun V, Ranftl R (2022) Language-driven
  semantic segmentation. In: International Conference on Learning
  Representations, \urlprefix\url{https://openreview.net/forum?id=RriDjddCLN}

\bibitem[{Shah et~al(2023)Shah, Osi{\'n}ski, Levine et~al}]{lmnav}
Shah D, Osi{\'n}ski B, Levine S, et~al (2023) Lm-nav: Robotic navigation with
  large pre-trained models of language, vision, and action. In: Conference on
  Robot Learning, PMLR, pp 492--504

\bibitem[{Huang et~al(2023)Huang, Mees, Zeng, and Burgard}]{vlmaps}
Huang C, Mees O, Zeng A, Burgard W (2023) Visual language maps for robot
  navigation. In: 2023 IEEE International Conference on Robotics and Automation
  (ICRA), IEEE, pp 10,608--10,615

\bibitem[{Radford et~al(2023)Radford, Kim, Xu, Brockman, McLeavey, and
  Sutskever}]{whisper}
Radford A, Kim JW, Xu T, Brockman G, McLeavey C, Sutskever I (2023) Robust
  speech recognition via large-scale weak supervision. In: International
  Conference on Machine Learning, PMLR, pp 28,492--28,518

\bibitem[{Wang and Komatsuzaki(2021)}]{gpt-j}
Wang B, Komatsuzaki A (2021) {GPT-J-6B: A 6 Billion Parameter Autoregressive
  Language Model}. \url{https://github.com/kingoflolz/mesh-transformer-jax}

\bibitem[{Black et~al(2022)Black, Biderman, Hallahan, Anthony, Gao, Golding,
  He, Leahy, McDonell, Phang, Pieler, Prashanth, Purohit, Reynolds, Tow, Wang,
  and Weinbach}]{gpt-neox20b}
Black S, Biderman S, Hallahan E, Anthony Q, Gao L, Golding L, He H, Leahy C,
  McDonell K, Phang J, Pieler M, Prashanth US, Purohit S, Reynolds L, Tow J,
  Wang B, Weinbach S (2022) Gpt-neox-20b: An open-source autoregressive
  language model. \eprint{2204.06745}

\bibitem[{Brown et~al(2020)Brown, Mann, Ryder, Subbiah, Kaplan, Dhariwal,
  Neelakantan, Shyam, Sastry, Askell, Agarwal, Herbert-Voss, Krueger, Henighan,
  Child, Ramesh, Ziegler, Wu, Winter, Hesse, Chen, Sigler, Litwin, Gray, Chess,
  Clark, Berner, McCandlish, Radford, Sutskever, and Amodei}]{gpt3}
Brown TB, Mann B, Ryder N, Subbiah M, Kaplan J, Dhariwal P, Neelakantan A,
  Shyam P, Sastry G, Askell A, Agarwal S, Herbert-Voss A, Krueger G, Henighan
  T, Child R, Ramesh A, Ziegler DM, Wu J, Winter C, Hesse C, Chen M, Sigler E,
  Litwin M, Gray S, Chess B, Clark J, Berner C, McCandlish S, Radford A,
  Sutskever I, Amodei D (2020) Language models are few-shot learners.
  \eprint{2005.14165}

\bibitem[{Verschueren et~al(2014)Verschueren, De~Bruyne, Zanon, Frasch, and
  Diehl}]{rcmodel}
Verschueren R, De~Bruyne S, Zanon M, Frasch JV, Diehl M (2014) Towards
  time-optimal race car driving using nonlinear mpc in real-time. In: 53rd IEEE
  conference on decision and control, IEEE, pp 2505--2510

\bibitem[{Lucia et~al(2017)Lucia, T{\u{a}}tulea-Codrean, Schoppmeyer, and
  Engell}]{MHE}
Lucia S, T{\u{a}}tulea-Codrean A, Schoppmeyer C, Engell S (2017) Rapid
  development of modular and sustainable nonlinear model predictive control
  solutions. Control Engineering Practice 60:51--62

\bibitem[{Jatavallabhula et~al(2023)Jatavallabhula, Kuwajerwala, Gu, Omama,
  Chen, Li, Iyer, Saryazdi, Keetha, Tewari et~al}]{conceptfusion}
Jatavallabhula KM, Kuwajerwala A, Gu Q, Omama M, Chen T, Li S, Iyer G, Saryazdi
  S, Keetha N, Tewari A, et~al (2023) Conceptfusion: Open-set multimodal 3d
  mapping. arXiv preprint arXiv:230207241

\bibitem[{Gresenz et~al(2021)Gresenz, White, and Schmidt}]{dataset1}
Gresenz G, White J, Schmidt DC (2021) Attention regions in terrain roughness
  classification for off-road autonomous vehicles. In: " "

\bibitem[{Dice(1945)}]{dice}
Dice LR (1945) Measures of the amount of ecologic association between species.
  Ecology 26(3):297--302

\end{thebibliography}
\endgroup






\end{document}